%% file: main.tex
\pdfoutput=1

\documentclass[11pt]{article}

\usepackage[]{ACL2023}

\usepackage{algorithm}
\usepackage{algpseudocode}

\usepackage{amsmath}
\usepackage{times}
\usepackage{mathtools}
\usepackage{amsmath}
\usepackage{amssymb}
\usepackage{bbm}
\usepackage{soul,color}
\usepackage{booktabs}
\usepackage{multirow}
\usepackage[capitalize]{cleveref}
\usepackage{hyperref}
\usepackage{adjustbox}
\usepackage{bm}
\usepackage{arydshln}
\usepackage{enumitem}

\usepackage{times}
\usepackage{latexsym}
\usepackage{amssymb}

\usepackage[T1]{fontenc}

\usepackage[utf8]{inputenc}

\usepackage{microtype}

\usepackage{inconsolata}

\newcommand\blfootnote[1]{%
  \begingroup
  \renewcommand\thefootnote{}\footnote{#1}%
  \addtocounter{footnote}{-1}%
  \endgroup
}

\newcommand\model{ReMask}

\title{\model{}: A Robust Information-Masking Approach for Domain Counterfactual Generation}
\author{Pengfei Hong$^{1*}$, Rishabh Bhardwaj$^{1*}$, Navonil Majumdar$^1$, Somak Aditya$^2$, Soujanya Poria$^1$ \\\\
$^1$ ISTD, Singapore University of Technology and Design, $^2$ Department of CSE, IIT Kharagpur\\
\texttt{\{pengfei\_hong,  navonil\_majumder\}@sutd.edu.sg}\\
\texttt{rishabh\_bhardwaj@mymail.sutd.edu.sg}\\
\texttt{saditya@cse.iitkgp.ac.in}, \texttt{sporia@sutd.edu.sg}
}

\begin{document}

\maketitle
\begin{abstract}

Domain shift\blfootnote{$^*$Equal Contribution} is a big challenge in NLP, thus, many approaches resort to learning domain-invariant features to mitigate the inference phase domain shift. Such methods, however, fail to leverage the domain-specific nuances relevant to the task at hand. To avoid such drawbacks, domain counterfactual generation aims to transform a text from the source domain to a given target domain. However, due to the limited availability of data, such frequency-based methods often miss and lead to some valid and spurious domain-token associations. Hence, we employ a three-step domain obfuscation approach that involves frequency and attention norm-based masking, to mask domain-specific cues, and unmasking to regain the domain generic context. Our experiments empirically show that the counterfactual samples sourced from our masked text lead to improved domain transfer on 10 out of 12 domain sentiment classification settings, with an average of 2\% accuracy improvement over the state-of-the-art for unsupervised domain adaptation (UDA). Further, our model outperforms the state-of-the-art by achieving 1.4\% average accuracy improvement in the adversarial domain adaptation (ADA) setting. Moreover, our model also shows its domain adaptation efficacy on a large multi-domain intent classification dataset where it attains state-of-the-art results. We release the codes publicly at \url{https://github.com/declare-lab/remask}.
\end{abstract}


\section{Introduction}

Despite significant advances in unsupervised representation learning, natural language processing (NLP) systems often strongly rely on expensive human-annotated datasets. These (labeled) datasets, however, are only available in specific domains. Systems trained on such datasets usually significantly under-perform on out-of-domain (OOD) samples during inference, due to strong reliance on the dataset-specific token- or feature-label correlations. These correlations do not often generalize beyond the domain of the training dataset. These correlations can even be spurious annotation artifacts. For example, in a sentiment-labeled dataset on restaurant reviews, the token \emph{food} may frequently appear in samples with negative sentiment score, due to selection bias~\cite{Veitch2021CounterfactualIT}. 
Such biases are particularly prevalent in low resource settings, where training instances are scarce~\cite{nan-etal-2021-uncovering}.


In response, annotating samples from new domains may prove to be expensive and infeasible in the long term. To address these issues, many domain adaptations (DA) techniques have been proposed~\cite{roark-bacchiani-2003-supervised, Daum2006DomainAF, BenDavid2010ATO, Jiang2007InstanceWF, Rush2012ImprovedPA, schnabel-schutze-2014-flors}. Many of such techniques~\cite{blitzer-etal-2007-biographies, Ziser2016NeuralSC, Ganin2015DomainAdversarialTO, BenDavid2020PERLPD} learn domain-invariant features that sidestep the pitfall of relying on domain-specific features when faced with OOD inputs. This obviously also hinders performance on in-domain samples where domain-specific features may be relevant. To address this issue, \citet{calderon2022docogen} recently proposed domain-counterfactual generation ({DoCoGen}) to transform given in-domain samples into out-domain samples. 

\begin{figure*}
    \centering
    \includegraphics[width=0.8\textwidth]{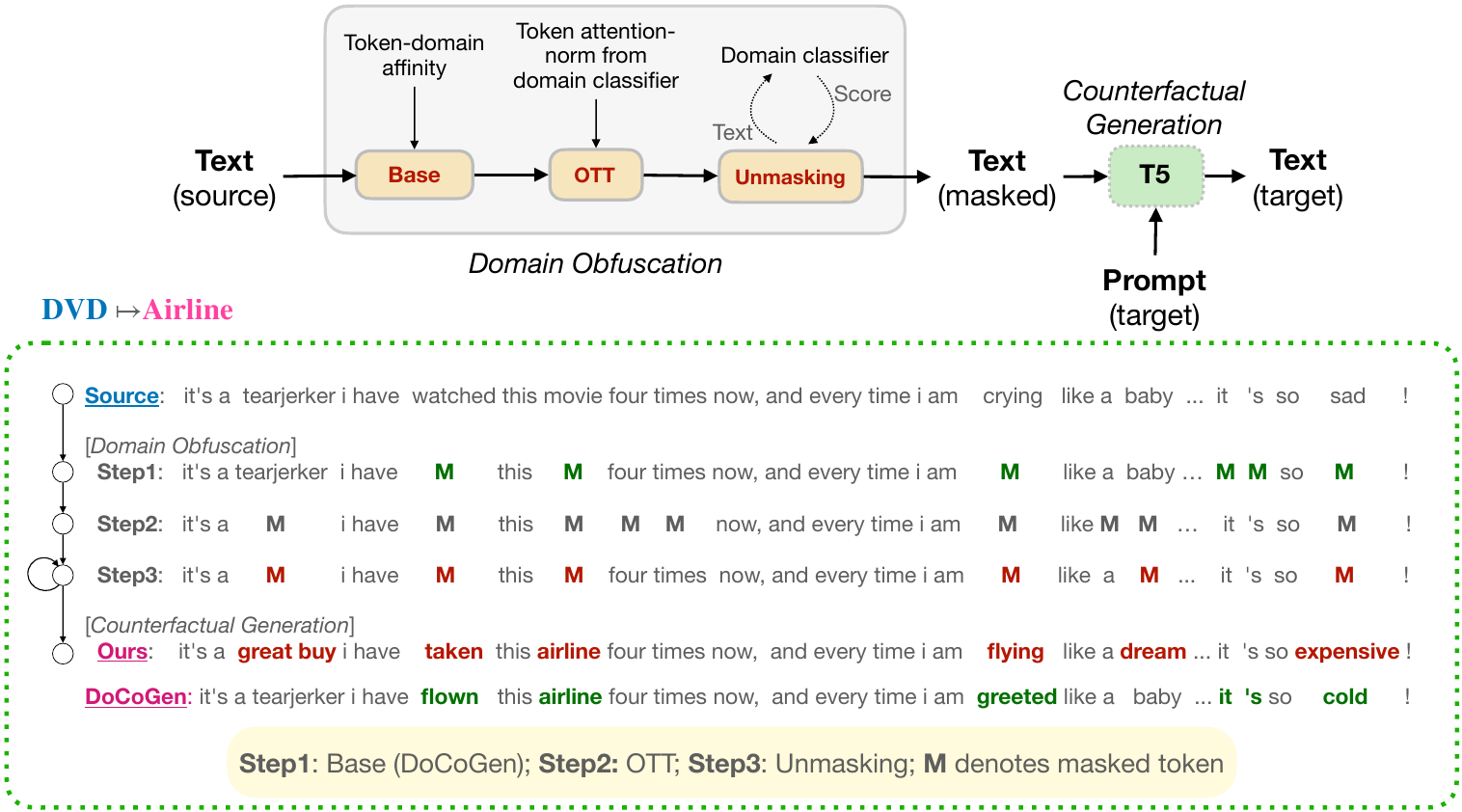}
    \caption{\footnotesize This diagram outlines the flow of our approach (top part) and illustrates it with an example (bottom part); the input {\bf Text} (source) is sequentially fed through \emph{Base}, \emph{OTT}, and \emph{Unmasking} steps to produce source-domain unaware masked text, which is passed to a \texttt{T5}-based generator to obtain domain counterfactual.}
    \label{fig:methodology}
\end{figure*}

{DoCoGen} masks the source-domain-specific n-grams in the input using a token-frequency-based approach. These masks are filled with target-domain-specific tokens using a conditional text generation language model that is fine-tuned with an unsupervised sentence reconstruction objective. However, the frequency-based masking approach does not account for the context to identify the source-domain-specific tokens. Moreover, this approach is limited by the statistics of an incomplete training set. These shortcomings may cause the approach to overlook many valid domain-token associations. For instance, in \cref{fig:methodology} the word \emph{tearjerker} is not masked by {DoCoGen}, where this word may be associated with the DVD of a tragedy movie. Furthermore, such frequency-based masking fails to generalize well to out-of-vocabulary words. 

Therefore, we propose a robust masking method that can achieve better retention of non-domain information, while removing domain-specific information. Our method includes three phases: frequency-based mask initialization~\cite{calderon2022docogen}, over-the-top (OOT) masking, and unmasking. We first phase is simply to initialize the mask using the frequency-based strategy by \citet{calderon2022docogen}. These masks are based on token-domain association consisting of two factors: the probability of the presence of a target token in a particular domain and the non-uniformity of this probability distribution. Tokens surpassing a certain association score, w.r.t. both source and target domain, are masked. In the second phase, we leverage the encoded knowledge in language models (LM)~\cite{Liu2019RoBERTaAR} to identify additional token-domain associations, improving recall at the risk of hurting precision. We found that a token with high attention-norm~\cite{kobayashi2020attention}, in a fine-tuned LM as domain classifier, could be strongly associated with the predicted domain. Such tokens with high enough attention norms are masked. Finally, we sequentially unmask the masked tokens, guided by the confidence score of a domain classifier. To minimize spurious token-domain associations from the prior two masking phases that hurt precision, the tokens that cause low domain confidence upon unmasking are kept unmasked.

The overall contributions of this work are: {\it i)} a robust masking strategy to mask more domain-specific tokens, as compared to the state of the art (SOTA)~\cite{calderon2022docogen}, while mitigating spuriously masked tokens; {\it ii)} domain counterfactuals (D-CON) from our model outperform the SOTA on binary and multi-label classification tasks, under unsupervised domain adaptation (UDA) and adversarial domain adaptation (ADA) settings, under almost all domain shifts. 

\section{Methodology}

\paragraph{Task Description.} Let $X_\mathcal{D}$ denote a text sampled from domain $\mathcal{D}$. We define a counterfactual generator function $f_c^{\mathcal{D}\rightarrow\mathcal{D}'}(X_{\mathcal{D}})$ as a mapping from a text $X_\mathcal{D} \sim \mathcal{D}$ to a text $X_\mathcal{D'} \in \mathcal{D}'$, such that, all but domain $\mathcal{D}$-specific information is preserved in $X_\mathcal{D}'$. We denote the domain-agnostic information by $X$ and the domain-specific text to be a specific fusion of $X$ and $\mathcal{D}$. We define a fusion function $\mathcal{I}$, such that, $X_{\mathcal{D}} \coloneqq \mathcal{I}(X;\mathcal{D})$. An inverse mapping $\mathcal{I}^{-1}$ makes it feasible to efficiently disentangle and extract the domain-generic information $X$. Therefore, the task of domain counterfactual generation aims to find a function that essentially disentangles the domain-generic information from $X_{\mathcal{D}}$ to obtain $X$, followed by mapping $X$ to a domain-specific text $X_{\mathcal{D}'}$. Thus, an ideal $f_c^{\mathcal{D}\rightarrow\mathcal{D}'}$ can be considered to be functionally equivalent to $\mathcal{I}(\mathcal{I}^{-1}(X_\mathcal{D}), \mathcal{D}')$. 

\paragraph{Overview.} The proposed approach consists of two phases:
\begin{itemize}[leftmargin=*, wide, nosep, labelwidth=!, labelindent=0pt]
    \item {\it Domain Obfuscation}: This phase masks a text $\mathcal{X}_D$, such that, it becomes void of source-domain-specific information. To this end, we propose a three-step masking approach.
    \item {\it Domain Counterfactual Generation}: As \citet{calderon2022docogen}, we feed the masked text to a T5-based encoder-decoder model $f_c$ to generate a counterfactual in the target domain $\mathcal{D}'$. 
\end{itemize}

\subsection{Domain Obfuscation}
We propose a novel three-step approach for domain corruption to mask more words carrying domain-specific cues and unmask words that are not specific to any domain. Masking words carrying domain-specific information helps the model learn to rely more on the input prompt, allowing us to have higher control over the generated text. Moreover, unmasking domain-generic words preserve the context information of text so as to keep the model-generated output an equivalent text to input but in the specified domain.

\paragraph{Step 1 (Base Masking).} We begin with the frequency-based (heuristic) masks, proposed by \citet{calderon2022docogen}. It assigns an affinity-based score to a word $w \in X$. The word ($w$)-domain($\mathcal{D}$) affinity is defined by
\begin{equation*}
    \rho(w, \mathcal{D})=P(\mathcal{D} \mid w) \cdot\left(1-\frac{H(D \mid w)}{\log N}\right),
\end{equation*}
where $D\in \{1,\ldots, N\}$ is a random variable representing a $N$ domain classes. $H(D|w)$ denotes the entropy of random variable $D|w$. A low value of $H(D|w)$ signifies that the spread of word $w$ is skewed towards fewer domains, and thus domain-specific. $P(\mathcal{D}{\mid}w)$ dictates the chances of domain being $\mathcal{D}$ given the word $w$. Thus, a word is said to have a high affinity with domain $\mathcal{D}$ if $\rho(w, \mathcal{D})$ returns a high value. To perform $X_\mathcal{D} \rightarrow X_\mathcal{D}'$ counterfactual generation, we mask a word $w \in X$ if the word has a high affinity towards $\mathcal{D}$, relative to $\mathcal{D}'$. Thus, the domain transfer (masking) score can be defined by

\begin{align*}
    \mathrm{m}_a\left(w, \mathcal{D}, \mathcal{D}^{\prime}\right){\coloneqq}\rho(w, \mathcal{D}){-}\rho\left(w, \mathcal{D}^{\prime}\right).
\end{align*}

The higher the value of $\mathrm{m}_a\left(w, \mathcal{D}, \mathcal{D}^{\prime}\right)$, the harder it is to perform its $\mathcal{D} \rightarrow \mathcal{D}'$ transfer. A word is masked if its $\mathrm{m}_a$-score is above a predefined threshold $\tau_1$. Generalizing the approach to n-grams, we first mask all the unigram words. This is followed by identifying the bigrams that do not overlap with the masked unigram, and then trigrams that do not overlap with the masked tokens\footnote{Threshold $\tau_1$ is a hyperparameter, best found to be 0.08}.

\paragraph{Step 2 (Over-The-Top Masking).} After heuristic masking, which is word context agnostic, we perform attention-based over-the-top (OTT) masking. The OTT masking serves two purposes---1) Context awareness: based on their context information, it reconsiders the words for masking that failed $m_a$-scoring and remained unmasked; and 2) Induction: it generalizes the notion of a domain to a more comprehensive set of word vocabulary and their word co-occurrences in which the domain-specific corpus is prone to be deficient. The context-aware attribute masks generic words, that under a given context and style of text, become domain informative. The inductive masking attribute makes use of the underlying language model to exploit the learned word-word co-occurrences to reconsider the unmasked words. Thus, considering words missing in the domain-specific corpus as well as words whose frequencies in the corpus do not truly reflect their domain affinity. This results in improved recall in domain-specific token extraction, although at the risk of hurting precision.

To perform OTT masking, we train a domain classifier $f_d: \mathcal{D} \rightarrow D$ that learns to classify an $X_\mathcal{D} \sim \mathcal{D}$ to a domain label $d \in D$. For finding OTT words to mask, we use an attention-based architecture as domain classifier $f_d$. Norm-based attention analysis has been seen to efficiently capture the importance of a word to the model's prediction \cite{kobayashi2020attention}. Following this observation, we define the attention-based domain affinity of a word as
\begin{align*}
    \mathrm{m}_b \left(w;l\right) \coloneqq ||\alpha_{w}^{l}\bm{v}_w^l||,
\end{align*}
where $||\cdot||$ computes Euclidean norm. In layer $l$ operations, let $\mathbf{W}_V^l$, $\bm{b}_V^l$ denote the value matrix and biases, $\mathbf{W}_O^l$ be the output projection. For a word $w$, let $\bm{y}_w^l$ be its representation at the output of layer $l$, we define $\bm{v}_w^l=(\bm{y}_w^{l-1} \mathbf{W}_{V}^l+\bm{b}_{V}^l)\mathbf{W}_{O}^l$, $\alpha_{w}^{l}$ specifies how much the classification token\footnote{Classification token for RoBERTa is \texttt{<s>}} in layer $l$ attends to $w$. We mask all the words whose $\mathrm{m}_b$ score is higher than a threshold denoted by $\tau_2$. $\tau_2$ is set as a hyperparameter based on each $X_{\mathcal{D}}$, this will be explained further in step 4.


\paragraph{Step 3 (Unmasking).} One of the aims of domain-counterfactual text generation is to preserve everything ---- including the task label --- except domain information. Therefore to make the minimal edits to its original text~\cite{calderon2022docogen}, we reduce the number of edits to by restoring the masked tokens from prior steps. Meanwhile, we posit that domain information masked by OTT (Step 2) is not unmasked in Step 3. The latter is controlled by thresholding the domain classification score of the masked sentence produced in this step 2. Analogous to counterfactually-augmented data (CAD) generation~\cite{Kaushik2020Learning}, which is done by minimally intervening on examples to change their ground truth label. In our case, we only want to use a minimal number of masks to remove the amount of domain-specific terms in the text. Specifically, we use the domain score given by the domain classifier to measure the amount of domain information contained in the example and make sure it is below a threshold~\footnote{In our initial experiments, we found the threshold value 0.4 works best. we also tried thresholding based on misclassification, in our observations, it tends to unmask a lot of context-specific words.}.

Additionally, Step 1 and Step 2 are prone to excessive masking from spurious domain-token correlations, resulting in the improved recall at the risk of poor precision. To account for these factors, we aim to unmask tokens that do not provide domain-specific cues, hence, restoring more contextual information; this is inspired by \citet{meng2022locating}, where \emph{causal tracing} is used to determine the association between different factors and the output in a large language model. Given input text $X_{\mathcal{D}}$, following Step 1 and Step 2, we obtain a masked text $\Tilde{X}^K_{\mathcal{D}}$, where $K=\{k_1, ...,k_n\}$ represents $n$ positions of the masked tokens. 
We define a subset of indices $U=\{k_i, ...,k_j\} \subseteq K$ containing the positions of the words to be unmasked. Thus, $\Tilde{X}^{K-U}$ represents the rectified masked output.


First, we independently intervene on the masked text $\Tilde{X}_{\mathcal{D}}$ by unmasking a token at position $i$ and perform domain classification using $f_d$, to obtain the change in domain-label probability of $D$.
We also define
\begin{equation*}
    \mathrm{m}_u(w_{k_i}) {\coloneqq} f_d^{\mathcal{D}}(\Tilde{X}_\mathcal{D}^{K-\{k_i\}}){-}f_d^{\mathcal{D}}(\Tilde{X}_\mathcal{D}),
\end{equation*}
where $f_d^{\mathcal{D}}(\cdot)$ returns a probability score of the origin domain $\mathcal{D}$. Per iteration, we sequentially restore the tokens in $K$, in the ascending order of $\mathrm{m}_u(w_{k_i})$. We stop the unmasking iteration when the condition $f_d^{\mathcal{D}}(\Tilde{X}_\mathcal{D}^{K-U}) < \tau_3$ is violated. Then we get the unmasked example. We posit that the domain specificity is highly correlated with $\mathrm{m}_u(w_{k_i})$. Therefore, we find taking the greedy approach to unmasking from the most domain-descriptive tokens robust in finding the optimal $U$.
%

\begin{table*}[]
\centering
\begin{adjustbox}{max width=\textwidth}
\begin{tabular}{@{}l@{}}
\bottomrule
\begin{tabular}[c]{@{}l@{}}\textcolor{purple}{Original} from \textbf{[dvd]}: Pippin dvd, I loved Ben Vereen in the show. Wanted the music for my ipod, too. Very satisfying! \end{tabular} \\\hline
\begin{tabular}[c]{@{}l@{}}\textcolor{purple}{Masking} [\textcolor{blue}{DoCoGen}] to \textbf{[book]}: Pipin <m>, I loved \textcolor{red}{Ben Vereen} in the <m>. Wanted <m> <m> for my \textcolor{red}{ipod} too. Very satisfying!
\end{tabular} \\
\begin{tabular}[c]{@{}l@{}}\textcolor{purple}{Masking} [\textcolor{blue}{Ours}] to \textbf{[book]}: Pipin <m>, I loved <m> <m> in the <m>. Wanted \textcolor{green}{the} for my <m> too. Very satisfying!
\end{tabular} \\ \bottomrule

\begin{tabular}[c]{@{}l@{}}\textcolor{purple}{Original} from \textbf{[electronics]}: sony rm - ax4000 this is a poorly designed remote. \\ i have six devices connected to my hd television set. the software depends on the user assigning positions to the various inputs to the tv and, \\in my case, routinely activated the wrong device when i used the remote. i purchased a logistics harmony remote and it works perfectly.
 \end{tabular} \\\hline
\begin{tabular}[c]{@{}l@{}}\textcolor{purple}{Masking} [\textcolor{blue}{DoCoGen}] to \textbf{[kitchen]}: <m> - \textcolor{red}{ax4000} this is a poorly <m> .  i have six <m> to my <m> set . the <m> depends on the \\ <m> assigning \textcolor{red}{positions} to the various <m> to <m> , in <m> routinely <m> wrong <m> when <m> the <m> . <m> logistics <m> and <m> . 
\end{tabular} 
\\
\begin{tabular}[c]{@{}l@{}}\textcolor{purple}{Masking} [\textcolor{blue}{Ours}] to \textbf{[kitchen]}: <m> - <m> this is a poorly \textcolor{green}{designed} <m> . i have six <m> to my <m> set . the <m> depends on the \\ <m> assigning <m> to the various <m> , \textcolor{green}{in my case} , routinely <m> the wrong <m> when i <m> . i purchased a <m> and \textcolor{green}{it works perfectly}.
\end{tabular} \\ \bottomrule
\end{tabular}
\end{adjustbox}
\caption{\footnotesize Domain Obfuscation (Masking).}
\label{tab:examples}
\end{table*}

\subsection{Domain Counterfactual Generation}

\paragraph{Training.} We do not have access to parallel domain counterfactuals. Thus, following \citet{calderon2022docogen}, we train a T5-based encoder-decoder model $M$ in an unsupervised manner. We essentially train the model to reconstruct the original text $X$ in the source domain from its masked form $\Tilde{X}$, which is presumably purged of source-domain-specific cues.
To have control on the domain of the generated text, we follow DoCoGen's approach of prepending a soft prompt $v$, indicating the domain, to the masked text in the input. The soft prompt is initialized using the embedding of the word\footnote{Following DoCoGen, we choose the most common word in that domain as the representative of that domain, we list the words we use in the appendix} representative of the target domain $\mathcal{D}$ and trained together with the text reconstruction objective. During training, the model learns to generate the original example, given the domain obfuscated text and the soft prompt: $M(\Tilde{X}, v) \rightarrow X$.

\paragraph{Inference.} 
Given $(x, \mathcal{D}, \mathcal{D}')$, we first feed through the $x$ through our domain obfuscation process to get $\Tilde{x}$ and select the soft prompt $v'$ to represent $\mathcal{D}'$. We feed $(\Tilde{x}, v')$ to the model to generate the domain counterfactual $x'$. We used beam search with a beam size of four for decoding $x'$.


\section{Intrinsic Evaluation}
Conditional generation is difficult to automatically evaluate. Therefore, we asked three Ph.D. students trained in natural language processing to manually evaluate the dataset considering the following evaluation measures: (1) Domain Relevance (D.REL) - whether the topic of the generated text is related to the target domain; (2) Label Preservation (L.PRES) - what is the label of the generated text and if the original label preserved; (3) Linguistic Acceptability (ACCPT) - how logical and grammatical the example is (on a 1-5 scale); and (4) Word Error Rate (WER) - what is the minimum number of word substitutions, deletions, and insertions required to make the example logical and grammatical. The test is conducted on 20 reviews, uniformly distributed among four domains (A, D, E, K) and 60 generated domain counterfactuals using DoCoGen and \model{}.

\cref{tab:intrinsic evaluation} shows that our masking achieves better scores than its baseline DoCoGen method, especially in preserving the original example label, it achieves a score of 95\%, surpassing its heuristic counterpart. We suspect \model{} keeps more information for the generation model to infer the label thus preserving label-invariance. We analyse the reason for this further in Section \ref{s:result}

\begin{table}[t]
\centering
\small
\resizebox{\linewidth}{!}{
    \begin{tabular}{@{}lcccc@{}}
    \toprule
    \textbf{Model} & \textbf{$\uparrow$ D.REL} & \textbf{$\uparrow$ L.PRES} & \textbf{$\uparrow$ ACCPT} & \textbf{$\downarrow$ WER} \\ \midrule
    DoCoGen          & 85    & 87.5  & 4.21 & 1.05 \\
    \model{}         & 92.5  & 95.0  & 4.33 & 1.0  \\
    Original Reviews & 100.0 & 100.0 & 4.94 & 0    \\ \bottomrule
    \end{tabular}
        }
    \caption{\footnotesize Human intrinsic evaluation. Up arrows ( $\uparrow$ ) represent metrics where higher scores are better, and down arrows $(\downarrow)$ represent the opposite.}
    \label{tab:intrinsic evaluation}
\end{table}

\section{Experimental Settings}

\subsection{Tasks and Settings}
We focused on two low-resource scenarios: unsupervised domain adaptation (UDA) and any domain adaptation (ADA). UDA assumes the availability of large unlabeled data from a source domain and target domain, as well as access to the limited number of labeled examples in the source domain. We follow \citet{calderon2022docogen} and choose to sample 100 labeled data in the source domain for both of the tasks mentioned below.  An even more challenging and potentially more realistic setting is ADA, which assumes no access to both labeled data and unlabeled data from the target domain in training time. In other words, the model will not have any information about the target domain distribution. Following a large body of DA work, we benchmark the proposed \model{} on cross-domain sentiment classification task and cross-domain multi-label intent classification task. For brevity, we refer readers to \citet{calderon2022docogen} for a detailed description of each dataset.

\paragraph{Sentiment Classification}
In this task, we combine 3 datasets together to form a dataset of 6 domains. The dataset includes 4 domains - Books(B), DVDs(D), Electronic items(E), and Kitchen Appliances(K) from product review multi-domain dataset~\cite{blitzer-etal-2007-biographies}; the challenging airline review dataset (A)~\cite{airline}; and the restaurant (R) domain from Yelp dataset challenge~\cite{wei-zou-2019-eda}.
We benchmark the sentiment classification task on UDA and ADA settings. In UDA, we focus on four of the six domains: A, D, E, and K. This will result in 12 (3$\times$4) cross-domain pairs. In this setting, the model can access all of the unlabeled source domain and target domain data. whereas For ADA, where an unlabeled dataset from the target domain is not within reach, our experiment uses A, D, E, and K as source domains; B, and R as target domains, thus resulting in a total of 8 (4$\times$2) domain pairs for ADA. Further, To facilitate the comparison with previous work, we focused on low resource settings by randomly sampling 100 labeled examples each from the four domains to as labeled source dataset. The reported scores are averaged across 5 training and development sets using different seeds.

\paragraph{Multi-Label Intent Prediction.}

The second task is to predict the potential intents of the utterances arising from information-seeking dialogs, as there could be multiple intents originating from the same utterances, it is treated as a multi-label classification problem. We choose the MANtIS dataset~\cite{Penha2019IntroducingMA}, inside each utterance could have eight potential intent labels. Following \citet{calderon2022docogen}, we only consider 5 most common labels: Further Detail (FD), potential answer (PA), Information Request (IR), Greetings and Gratitude (GG), Original Question (OQ).\\
The MANtIS dataset is a multi-domain dataset with 14 domains. We experiment with the UDA setting. To ease comparison with {DoCoGen}, we choose the first six domains: Apple (AP), DBA (DB), Electronics (EL), Physics(PH), Statistics (ST), askubuntu (UB) with available unlabeled data to form 30 (5$\times$6) cross-domain pairs for UDA.

\input{reviews-uda-result.tex}
\input{reviews-ada-result.tex}
\input{mantis-result.tex}

\section{Models and Baselines} \label{s:baseline}

We tested on three types of models~\footnote{Experiment 3 is not applicable to the ADA setting due to lack of labeled data.}: (a) baseline models (b) variants for each step in our masking approach c) an upper-bound model to approximate the best performance using domain counterfactuals (D-CON) on the downstream tasks. Unless otherwise stated, all the domain classifiers and sentiment classifiers use the same model, based on a pretrained Roberta-base model, and all generation models are based on a pretrained \texttt{T5-base}.
\paragraph{Baseline DA Models.}
We experiment with four baselines: (1) No-Domain-Adaptation ({NoDA}), the model only trained on available training data from the source domain; (2) Random-masking Random-Reconstruction ({RMRR})~\cite{ng-etal-2020-ssmba}, randomly masks token from the input example and then fills the masks by masked language modeling head. 3) PERL~\cite{BenDavid2020PERLPD}, a SOTA for the UDA setup.  4) {DoCoGen}: Use the same model and training strategy as we do, but only use the first step of our masking procedure.
\paragraph{Ablations.}
We consider three types of ablations in our masking procedure. First, we show ablations for step 2 in our procedure. (5) 2-Attention-Score-Masking (2-ASM) where we use attention score rather than attention norm to perform OOT masking step, 
(6) 3WO where we show the performance of unmasking by following word order in the sentence. And finally, we show  (7) NoInit, where we remove the heuristic masking step, and only use OOT masking with unmasking. (8) No-Unmask, where we only have step 1,2 and removed step 3 unmasking.
\paragraph{Upper-Bound.}  To approximate the upper-bound for D-CON augmentation, we use \textit{Oracle-Matching} ({Oracle}) which can access to target domain labeled data~\cite{calderon2022docogen}. Given an example from a source domain, {Oracle} looks for the most similar example with the same label in the target domain as training data. 

\section{Results and Discussions} \label{s:result}
Tables \ref{tab:sentiment_uda} and \ref{tab:sentiment_ada}  present sentiment classification accuracy results for the 12 UDA and 8 ADA setups respectively. Table \ref{tab:mantis} presents the average intent prediction F1 scores for each source domain, taken across 5 target domains for the UDA setup. 

\paragraph{Results Comparison.}
For sentiment classification, our method \model{}, outperforms all baseline models in 10 of 12 UDA setups and in 6 of 8 ADA setups, gaining an average of 2\% and 1.4\% over the baseline masking approach in UDA and ADA settings, respectively. Furthermore, in two ADA setups on sentiment classification, our model outperforms the Oracle-Gen scores which suggests our model can effectively remove domain-specific information in these domain pairs. For intent prediction \footnote{lack of working code implementation and specifications to reproduce the reported results of \texttt{DoCoGen}}, our model outperforms four out of six setups, reaching an average gain of 0.8\% across domains. Moreover, NoInit result demonstrates that our model performance degrades if there is no initial mask provided by step 1. One key observation on the masking process is that Step 2 of our method fails to identify some domain-specific tokens in the presence of too many such tokens in the text. We suspect that the thresholding attention norm is not robust enough when the attention norm might be thinly spread among many domain-specific tokens. On the other hand, \model{} outperforms DoCoGen by 0.8\% on the MANtIS dataset under the UDA setup. This further confirms the efficacy of our method. We found that our \model{} have less performance improvement for domains where the domain specific words are distinctive like UB and have higher performance boost in domains that have more complex text, e.g. EL and PH.

\paragraph{Discussion on \model{}'s ability to retain contextual information and remove domain-specific information.}
To test our proposed method's effectiveness in removing domain-specific information, we train a domain classifier (based on the Roberta-base model) to predict the domain from the masked text. Table \ref{tab:domain_cls_accuracy} shows that the accuracy of the domain classifier trained on text masked by \model{} is consistently lower than \texttt{DoCoGen}, which proves the efficiency of our model ability in the obfuscation of the domain. The difference shrinks as the number of training examples increases. This may suggest that the model is still able to exploit the spurious correlation between linguistic features and the domain of the dataset.

\begin{table}[t]
\centering
\begin{tabular}{@{}lccc@{}}
\toprule
\textbf{Model / \#Train Samples} & \textbf{400} & \textbf{1k} & \textbf{10k} \\ \midrule
DoCoGen                & 40.6         & 70.4        & 86.6         \\
\model{}               & 15.4         & 44.1        & 85.2         \\ \bottomrule
\end{tabular}
\caption{Domain classification accuracy on the masked text between \model{} vs DoCoGen masking.}
\label{tab:domain_cls_accuracy}
\end{table}

\begin{table*}[t]
    \centering
    \small
    \resizebox{0.9\linewidth}{!}{
    \begin{tabular}{l|ccc|ccc|ccc|ccc|}
    \toprule
    \multirow{2}{*}{{\textbf{Domain}}}
    & \multicolumn{3}{c}{\textbf{A}}
    & \multicolumn{3}{c}{\textbf{D}}
    & \multicolumn{3}{c}{\textbf{E}}
    & \multicolumn{3}{c}{\textbf{K}}\\
    \cmidrule(lr){2-4}
    \cmidrule(lr){5-7}
    \cmidrule(lr){8-10}
    \cmidrule(lr){11-13}
    & Step1 & +Step2 & +Step3
    & Step1 & +Step2 & +Step3
    & Step1 & +Step2 & +Step3
    & Step1 & +Step2 & +Step3 \\
    \midrule
    A  & 15.2 & 27.8 & 16.3 & 37.9 & 48.16 & 44.5 & 37.3 & 54.2 & 30.9 & 38.0 & 56.1 & 33.3 \\
    D & 25.0 & 34.1 & 31.3 & 16.5 & 27.6 & 24.0 & 24.0 & 44.5 & 23.2 & 23.9 & 45.2 & 26.1  \\
    E  & 27.8 & 33.4 & 26.5 & 26.7 & 30.1 & 24.5 & 15.7 & 30.2 & 16.4 & 19.7 & 22.0 & 19.6  \\
    K  & 30.2 & 40.2 & 38.7 & 28.7 & 45.8 & 35.6 & 21.1 & 33.5 & 21.4 & 15.7 & 27.0 & 22.8  \\
    \bottomrule
    \end{tabular}
    }
    \caption{The average number of masks in each step.}
    \label{tab:masking_stat_avg}
\end{table*}

\paragraph{The Need of Step 2 and Step 3.}
As presented in \cref{tab:sentiment_uda}, \model{} without Step 3 (NoUnmask) performs very poorly, worse than DoCoGen. This might give us the impression that Step 2 is unnecessary. However, we hypothesize that Step 1 alone is not enough to correctly distinguish all the domain-indicative tokens --- as Step 1 is not context-sensitive. Step 2 addresses this issue which essentially aims at increasing the recall of the token-level masking. Although Step 2 can increase recall of toke-level masking, it has no control over the precision and hence can produce imperfect masks that are domain neutral or key to preserving contextual cues. One should also note that counterfactual generators should try to generate counterfactuals with minimal edits. So, we try to spot and remove these imperfect masks by a greedy unmasking strategy in Step 3. As such, Step 3 aims to produce masks without losing recall in Step 2 (OTT), but to improve the precision of Step 2-produced masks significantly.
We empirically verify the functioning of these steps in \cref{tab:masking_stat_avg}. We can see that the number of masks after Step 2 increases significantly, thus risking low precision and high recall (comparable to information retrieval problems). Step 3 unmasks such imperfect token-level masks resulting in fewer token-level masks than Step 2 in \cref{tab:masking_stat_avg}. We also show qualitative examples in \cref{tab:mask-unmask} to depict the usefulness of Step 2 and Step 3, where Step 3 unmasks critical domain neutral and contextual cue-bearing tokens as well as correctly retains important domain-dependant masks. This, as a result, maintains the recall of step 2 but increases its precision. Although we do not have a direct measurement of this precision and recall hypothesis due to the lack of ground truth labels of these masks --- we indirectly prove so using our experiments in \cref{tab:sentiment_uda}. The number of masks after Step 3 lies between Step 1 and Step 2 indicating Step 1 may not capture all the relevant masks and Step 2 might be excessively masking. If we further link this observation with \cref{tab:sentiment_uda}, we can conclude that Step 3 is required to improve performance over Step 1 (DoCoGen).

\begin{table}[t]
    \centering
    \resizebox{\linewidth}{!}{
    \begin{tabular}{p{12cm}}
    \hline
\textcolor{purple}{Source domain:} dvd; \textcolor{purple}{destination domain:} electronics \\
\textcolor{purple}{Original Text:} Pippin dvd, I loved Ben Vereen in the show. Wanted the music for my ipod, too. Very satisfying! \\
\textcolor{purple}{First Step Maksing:} Pipin <m>, I loved Ben Vereen in the <m>. Wanted the <m> for my ipod, too. Very satisfying! \\
\textcolor{purple}{Second Step OTT Masking:} \colorbox{red}{<m>} <m>, I loved \colorbox{green}{<m>} in the <m>. Wanted the <m> for my ipod, too. Very \colorbox{red}{<m>}! \\
\textcolor{purple}{Third Step Unmasking:} Pipin <m>, I loved <m> in <m>. Wanted the <m> for my ipod too. Very satisfying!\\
Generation: Pipin is great, I loved green color in the earphone. Wanted the earphones for my ipod too. Very satisfying!\\
\hline
\textcolor{purple}{Source domain:} airline, \textcolor{purple}{destination domain:} kitchen\\
\textcolor{purple}{Original Text:} 40 am departure time became 4. 30 am due to a crew issue. the plane was an older 737. the empower jacks at our seats were dead and there was grime on the hard part of the seat. there was no special meal service for business class - everybody got a small ham and cheese sandwich. the front toilet quickly got foul. \\
\textcolor{purple}{First Step Heuristic Masking:} \\
<m> time became 4 <m> to a <m> issue . the <m> . the empower jacks at our <m> dead and <m> grime on the hard part of the <m> . <m> special <m> for <m> - everybody got a small <m> and <m> . the <m> quickly got foul. \\
\textcolor{purple}{Second Step OTT Masking:} \\
<m> \colorbox{green}{<m>} became 4 <m> to a <m> \colorbox{red}{<m>}. the <m> \colorbox{red}{<m>} <m> . the \colorbox{green}{<m>} at our <m> \colorbox{red}{<m>} and <m> grime on the hard part of the <m> . <m> special <m> for <m> - \colorbox{green}{<m>} got a \colorbox{green}{<m>} <m> and <m> . the <m> quickly got \colorbox{green}{<m>}. \\
\textcolor{purple}{Third Step Unmasking:} \\
<m> <m> became 4 <m> to a <m> issue. the <m> was <m> . the <m> at our dead and <m> grime on the hard part of the <m> . <m> special <m> for <m> - <m> got a <m> and <m> . the <m> quickly got <m>. \\
\hline
    \end{tabular}
    }
    \caption{\footnotesize A few examples of the masks produced by different steps of our approach; the masks highlighted with \colorbox{red}{<m>} are masks added in Step 2, but removed(unmasked) in Step 3, whereas \colorbox{green}{<m>} remains masked after Step 3; consecutive masks are at times merged into one mask for brevity.}
    \label{tab:mask-unmask}
\end{table}






\section{Related Work} \label{s:related-works}
Our work takes inspiration from several established lines of research, namely Domain Adaptation, Counterfactual Data augmentation, and counterfactual text generation.

\textbf{Domain Adaptation.} Domain Adaptation (DA; \cite{farahani2021brief}) deviates from the assumption that test data comes from the same distribution as training and is aimed at improving the performance of models in target domains (possibly) with a different distribution than the training source domain. In NLP, various DA setups are considered 
Approaches in Unsupervised Domain Adaptation \cite{Blitzer2006DomainAW} assumes the availability of unlabeled data from both source and target domains, as well as the existence of labeled data in the source domain.  In adversarial domain adaptation \cite{DBLP:conf/emnlp/WangGLLGW19}, the assumptions are the same as UDA, but adversarial learning is employed to learn domain-invariant representations. In contrast, and domain adaptation (ADA; \cite{DBLP:journals/tacl/Ben-DavidOR22}) approaches assumes no knowledge of the target domains at training time. We consider UDA and Adversarial DA settings in our work.

\textbf{Counterfactual Data Augmentation.} Data augmentation, in general, is an important sub-field of NLP that aims to mitigate problems introduced by the low-resource availability of text data. It aims to increase the number of examples for training without any explicit efforts to collect new samples. There have been rule-based approaches that work by modifying the underlying text by using some pre-defined heuristics \cite{wei-zou-2019-eda}. Beyond significant heuristic efforts, there have been model-based approaches to modify the words or generate prior distribution-based completely new samples \cite{kobayashi2018contextual}. Counterfactual Data Augmentation is a more sophisticated approach that performs a minimal text intervention (such as a specific concept), thus constructing an example with a modified label \cite{Kaushik2020Learning}.

\textbf{(Counterfactual) Controlled Text Generation.} 
In conditioned or controllable text generation \cite{prabhumoye-etal-2020-exploring}, the task is to generate text that satisfies certain pre-specified conditions (such as topic, sentiment or domain). Its useful for a vast range of applications including data augmentation and domain adaptation. Previous approaches involved finetuning the LM outputs using re-inforcement learning \cite{DBLP:journals/corr/abs-1909-08593}, training Generative Adversarial Networks (Yu et al., 2017), or training conditional generative models. Recent approaches adapted causal text modeling \cite{DBLP:journals/tacl/FederKMPSWEGRRS22} and uses counterfactual editing of text towards controlled text generation \cite{DBLP:conf/aaai/MadaanPPS21,DBLP:conf/acl/WuRHW20}. Most of these models experiment with shorter text and properties that are easier to \textit{ground} in the text (more specifically replacing one or two spans suffice).
\section{Conclusion}

This work empirically shows that the addition of attention norm-based masking leads to additional domain-token associations missed by the SOTA masking strategy. Furthermore, our iterative unmasking approach leads to the removal of spurious domain-token associations, resulting in improved domain obfuscation. These two innovations collectively allow better domain transfer on sentiment and intent classifications tasks.

\section*{Limitations}
We discuss the limitations of our work:
\begin{itemize}
    \item While the three-step masking is shown to be beneficial, masking (base+OTT) followed by unmasking may introduce several redundant computations as a token masked in the first two steps might get unmasked in step 3.
    \item When compared against the DoCoGen baseline \cite{calderon2022docogen}, iterative masking steps increase the time complexity of the domain obfuscation which leads to masking latency. Moreover, as the domain classifier is a critical part of the domain obfuscation, the approach has extra memory footprints.
    \item The proposed masking approach introduces two extra hyperparameters $\tau_2$ and $\tau_3$ on top of the hyperparameters introduced by DoCoGen. While we identify them as a fixed scalar value working for all kinds of input, we posit that one can propose dynamic input or source domain adaptive thresholding. Currently, we classify it as a limitation of the proposed work.
\end{itemize}

\section*{Ethics Statement}

For intrinsic evaluation, we engage three Ph.D. students who are fairly compensated.  This qualitative evaluation project passed ethics
review of our IRB as it does not
contain any confidential data.

\bibliographystyle{acl_natbib}
\bibliography{custom}

\appendix
\section{Appendix}

\subsection{Hyperparameters and Setups}
\paragraph{Data Preprocessing} We follow previous work in data preprocessing and truncate each example to 96 tokens, using the HuggingFace T5-base tokenizer. The hyper-parameter was set to 96 due to computation reasons and since the median number of words in the labeled examples was 89. When an example is longer than 96 tokens, we keep the ﬁrst 96 tokens. For example from the Airline domain, before truncating, we remove the ﬁrst sentence since it mostly contains details about the ﬂight (like “from JPK to LAX”).

\paragraph{Heuristic Masking}
We estimate $P(\mathcal{D} | w)$ for uni-grams, bi-grams and tri-grams which appear in the unlabeled data in at least 10 examples. We use the NLTK Snowball stemmer to stem each word of the n-grams. The smoothing hyperparameters in the computation of $P(\mathcal{D} | w)$ are set to be 1, 5, and 7 for uni-grams, bi-grams and tri-grams, respectively. We use a $\tau = 0.08$ threshold and mask an additional 5\% of the training examples (in order to add noise between training epochs). We set $\tau = 0.08$ since it resulted in the successful domain alternation of more than 80\% examples. 

\paragraph{Step 2 OTT}
The domain classifier to get a domain-related score is based on the pretrained HuggingFace distilroberta model. It has total of 6 layers and we use the attention norm in the middle layer (layer 3, 4) to  perform Over-The-Top masking

\paragraph{Domain Specific Prompt}
We choose four words representing the domain and initialize the domain-specific prompt with their word embeddings. The following is the word we use for each domain:
Airline: airline, ﬂight, seat, staff; DVD: dvd, character, actor, plot; Electronics: electronics, ipod, router, software; Kitchen: kitchen, dishwasher, pan, oven; and for the MANtIS dataset: Apple: apple, itunes, iphone, nacbook; askubuntu: askubuntu, ubuntu, apt, deb; DBA: dba, database, SQL, query; Electronics: electronics, schematics, voltage, circuit; Physics: physics, gravity, particle, quantom; Statistics: stats, regression, logits, variance;

\paragraph{Training of D-CONs generation model}
The controllable model is based on a pre-trained HuggingFace T5-base model. We train it on the unlabeled data for 20 epochs and pick the model whose generated examples for an unlabeled held-out set are of the highest domain accuracy (D.REL). 5 Training is performed with the AdamW optimizer learning rate parameter of 5e-5 and a weight decay parameter of 1e-5.

\paragraph{Task Classifiers}
Task classifiers are based on the Huggingface's RobertaForSequenceClassification model. We train the classiﬁers for 5 epochs with a batch size of 16 and pick the best model based on the performance on the validation set. Training is performed using the AdamW optimizer with learning rate parameters of 5e-5 for the encoder blocks and of 5e-4 for the linear layer and weight decay parameter of 1e-5.

\subsection{Samples}

In this section, we include masked text from each stage of our mask for our generated D-CONS. \\
(1) \underline{Original, \textbf{Airline}} \\
Was traveing with my partner, we got bored halfway through the flight as there is no inflight entertainment system . hot food was served, overall a positive experience . \\
\underline{Step 1, \textbf{Airline $\rightarrow$ Electronics}} \\
Was traveing with my partner, <m> halfway through the <m> is no <m> . <m> was <m> , <m> positive experience . \\
\underline{Step 2, \textbf{Airline $\rightarrow$ Electronics}} \\
Was <m> with my <m>, <m> through the <m> is no <m> . <m> was <m> , <m> positive <m> . \\
\underline{Step 3, \textbf{Airline $\rightarrow$ Electronics}} \\
Was <m> with my partner, <m> through the <m> is no <m> system. <m> was <m> , overall a positive experience . \\
\\
(2) \underline{Original, \textbf{Airline}} \\
both legs departed and arrived in time . approx 75 \% full with a mix of business leisure and Disney visitors . buy on board food is reasonably priced and the pre - order Irish breakfast was delicious . fare was just €100 return . recommended . \\
\underline{Step 1, \textbf{Airline $\rightarrow$ kitchen}} \\
both legs <m> and <m> in time . <m> 75 <m> a mix of <m> leisure and Disney visitors .  buy <m> is reasonably priced and the <m> irish <m> was <m> . <m> was just <m> return . <m> \\
\underline{Step 2, \textbf{Airline $\rightarrow$ kitchen}} \\
both <m> and <m> . <m> 75 <m> a mix of <m> and <m> .  buy on <m> is reasonably priced and the <m> was <m> . <m> was just <m> return . <m> \\
\underline{Step 3, \textbf{Airline $\rightarrow$ kitchen}} \\
both <m> and <m> . <m> 75 \% <m> a mix of <m> and <m> .  buy on <m> is resonably priced and the pre - order Irish <m> was <m> . <m> was just <m> return . recommended \\
\\
(3) \underline{Original, \textbf{Dvd}} \\
Startrek Voyager : I am a very avid " star trek " fan , and find the dvd ' s very worthwhile and interesting . ( my star trek interests go back to the original series with william shatner, leonard nimoy, james doohan, et. al. \\
\underline{Step 1, \textbf{Dvd $\rightarrow$ Electronics}} \\
<m> Voyager : I am a very avid <m> " <m>, and find the <m> ' s very worthwhile and <m> . (my star <m> go back to <m> with <m> , <m> nimoy , <m> doohan <m> . <m> . \\
\underline{Step 2, \textbf{Dvd $\rightarrow$ Electronics}} \\
<m> : I am a very avid <m>, and <m> 's <m> worthwhile and <m> . (my <m> go back to <m> .  \\
<m> : I am a very avid `` <m> `` <m> , and <m> ' s <m> worthwhile and interesting . ( my <m> go back to <m> . \\
\\
(4) \underline{Original, \textbf{Dvd}} \\
Tell all your friends! This is one of my favorite movies and i find it unfortunate that not too many people
  know about it. i think it is important for people to know this story. Everyone should watch this movie simply because it is fantastic . \\
\underline{Step 1, \textbf{Dvd $\rightarrow$ Kitchen}}
Tell all your friends! This <m> of my <m> and i find it unfortunate that not too many people
  know about it . i think it is important for people to know this <m> . <m> this <m> simply because it is fantastic . \\
\underline{Step 2, \textbf{Dvd $\rightarrow$ Kitchen}}
<m> friends ! This <m> of my <m> and i <m> unfortunate that not too many <m> it. i think it is important for <m> . <m> this <m> because it is <m> . \\
\underline{Step 3, \textbf{Dvd $\rightarrow$ Kitchen}}
Tell all your friends! This is one of my favorite <m> and i find it unfortunate that not too many people
  know about it. i think it is important for people to know this <m>. Everyone should <m> this <m> simply because it is fantastic . \\
  \\
(5) \underline{Original, \textbf{Electronics}} \\
Great service, cartridge like a new one ordering was fast and easy . could not detect any inferiority in the cartridges and will order them again \\
\underline{Step 1, \textbf{Electronics $\rightarrow$ Airline} } \\
Great service , cartridge like a <m> ordering was fast and easy . could not <m> any inferiority in the cartridges and will order them again . \\
\underline{Step 2, \textbf{Electronics $\rightarrow$ Airline}} \\
Great <m> , <m> like <m> was fast and <m> . could not <m> any <m> in the <m> and will <m> again .\\
\underline{Step 3, \textbf{Electronics $\rightarrow$ Airline}} \\
Great <m>, <m> like a new one <m> was fast and easy . could not <m> any inferiority in the <m> and will order them again \\
\\
(6) \underline{Original, \textbf{Kitchen}} \\
Conducts heat really well this set conducts heat really well and everything heats up fast but be careful not to burn anything which i did the first time using it . also it takes a bit of scrubbing if you burn it bt other than that it's a great product \\
\underline{Step 1, \textbf{Kitchen $\rightarrow$ Electronics}} \\
conducts <M> really well this <M> conducts <M> really well and everything <M> up fast but be careful not to     burn anything which i did the first time using it . also it takes a bit of <M> if you burn it bt other than that it ' s a great <M> \\
\underline{Step 2, \textbf{Kitchen $\rightarrow$ Electronics}} \\
<M> really well this <M> really well and everything <M> up fast but <m> to <m> anything which i did the first time <m> . also it takes a bit of <M> if you <m> it <m> other than that it ' s a great <M> \\
\underline{Step 3, \textbf{Kitchen $\rightarrow$ Electronics}} \\
<M> really well this <M> really well and everything <M> up fast but <m> to <m> anything which i did the first time using it . also it takes a bit of <M> if you <m> it <m> other than that it ' s a great product \\

\end{document}

%% file: reviews-uda-result.tex
\begin{table*}[t]
\centering
\begin{adjustbox}{max width=\textwidth}
\begin{tabular}{@{}lccccccccccccc@{}}
\toprule
\multicolumn{14}{c}{Source Domain $\rightarrow$ Target Domain ($\mathcal{D} \rightarrow \mathcal{D}'$)} \\ \midrule
\textbf{Model} &
  $\mathbf{A} {\rightarrow} \mathbf{D}$ &
  $\mathbf{A} {\rightarrow} \mathbf{E}$ &
  $\mathbf{A} {\rightarrow} \mathbf{K}$ &
  $\mathbf{D} {\rightarrow} \mathbf{A}$ &
  $\mathbf{D} {\rightarrow} \mathbf{E}$ &
  $\mathbf{D} {\rightarrow} \mathbf{K}$ &
  $\mathbf{E} {\rightarrow} \mathbf{A}$ &
  $\mathbf{E} {\rightarrow} \mathbf{D}$ &
  $\mathbf{E} {\rightarrow} \mathbf{K}$ &
  $\mathbf{K} {\rightarrow} \mathbf{A}$ &
  $\mathbf{K} {\rightarrow} \mathbf{D}$ &
  $\mathbf{K} {\rightarrow} \mathbf{E}$ &
  $\mathbf{A V G}$ \\
  \midrule
NoDA &
  69.4 &
  78.6 &
  78.2 &
  72.3 &
  80.2 &
  82.4 &
  81.0 &
  79.8 &
  87.6 &
  72.5 &
  78.6 &
  85.4 &
  78.8 \\
RM-RR &
  69.5 &
  80.1 &
  80.0 &
  72.3 &
  81.0 &
  83.8 &
  79.6 &
  79.5 &
  88.4 &
  70.6 &
  79.1 &
  84.5 &
  79.0 \\
 PERL &
 72.9 &
 81.1 &
 83.6 &
 81.5 &
 83.0 &
 86.9 &
 81.1 &
 81.7 &
 88.5 &
 77.9 &
 78.2 &
 86.1 &
 81.9 \\
DoCoGen &
  70.6 &
  79.7 &
  79.8 &
  75.8 &
  82.8 &
  84.4 &
  83.0 &
  82.0 &
  89.3 &
  81.2 &
  82.2 &
  87.3 &
  81.5 \\ 
DoCoGen-Rob &
  62.7&
  73.9&
  74.9&
  76.9&
  85.7&
  87.7&
  82.7&
  82.1&
  90.3&
  81.9&
  83.0&
  88.9&
  80.9 \\

\cmidrule(l){2-14} 

2-ASM &
68.7 &
81.8 &
79.5 &
81.0 &
88.3 &
88.9 &
82.0 &
84.4 &
91.4 &
80.7 &
80.9 &
88.1 &
82.9 \\
3WO &
65.4 &
78.4 &
81.5 &
80.3 &
81.0 &
80.0 &
85.4 &
75.6 &
85.2 &
84.2 &
80.0 &
82.1 &
79.9 \\
NoInit &
62.7 & 
73.9 &
74.9 &
76.9 &
85.7 &
87.7 &
82.7 &
82.1 &
90.3 &
81.9 &
82.95 &
88.9 &
80.9 \\
NoUnmask &
69.1 &
77.2 &
71.1 &
78.5 &
79.6 &
78.9 &
76.4 &
80.0 &
83.4 &
66.4 &
74.1 &
82.0 &
76.4 \\
NoOTT &
62.7 &
73.9 &
74.9 &
76.9 & 
85.7 &
82.7 &
82.1 &
90.3 &
81.9 &
83.0 &
88.9 &
80.9 &
80.8 \\
OUR &
  \textbf{74.4} &
  \textbf{85.3} &
  78.5 &
  \textbf{78.2} &
  \textbf{88.5} &
  \textbf{88.9} &
  81.0 &
  \textbf{82.9} &
  \textbf{89.7} &
  \textbf{81.5} &
  \textbf{83.4} &
  \textbf{89.7} &
  \textbf{83.5} \\ \cmidrule(l){2-14} 
Oracle-Gen &
  83.8 &
  88.4 &
  88.9 &
  83.6 &
  89.3 &
  90.0 &
  84.9 &
  84.6 &
  90.7 &
  84.1 &
  82.2 &
  89.0 &
  86.6 \\ \bottomrule
\end{tabular}
\end{adjustbox}
\caption{\footnotesize Unsupervised Domain Adaptation results for sentiment classification in the \cite{Blitzer2006DomainAW} dataset. DoCoGen-Rob uses a RoBERTa-base task classifier instead of T5 for a fair comparison to our model.}
\label{tab:sentiment_uda}
\end{table*}

%% file: reviews-ada-result.tex
\begin{table*}[t]
    \centering
    \resizebox{0.8\linewidth}{!}{
    \begin{tabular}{@{}lccccccccc@{}}
    \toprule
    Source~$\xrightarrow{}$ 
    &
    \multicolumn{2}{c}{$\mathbf{A}$} 
    &
    \multicolumn{2}{c}{$\mathbf{D}$}
    &
    \multicolumn{2}{c}{$\mathbf{E}$}
    &
    \multicolumn{2}{c}{$\mathbf{K}$}
    & \multirow{2}{*}{\textit{\textbf{AVG}}}\\
    \cmidrule{1-9}
    
    Target~$\xrightarrow{}$  & $\mathbf{B}$ & $\mathbf{R}$ & $\mathbf{B}$ & $\mathbf{R}$ & $\mathbf{B}$ & $\mathbf{R}$ & $\mathbf{B}$ & $\mathbf{R}$ &  \\
    \midrule

    NoDA &             69.1 & 76.5 & 82.3 & 82.8 & 81.5 & 84.5 & 82.4 & 85.2 & 80.5 \\
    RM-RR &            69.4 & 78.4 & 83.8 & 83.5 & 81.9 & 85.6 & 83.7 & 85.4 & 81.5 \\
    DoCoGen &          70.9 & 78.1 & 84.4 & 82.9 & 83.9 & 86.0 & 84.5 & 85.7 & 82.1 \\
    DoCoGen-Rob &      70.4 & 82.1 & 79.5 & 82.4 & 82.9 & 86.6 & 85.5 & 86.7 & 82.0 \\
    \midrule
    2-ASM &            65.7 & 76.3 & 87.1 & 82.3 & 80.8 & 84.9 & 85.0 & 82.3 & 80.7 \\
    NoInit &           71.4 & 82.4 & 87.3 & 82.2 & 84.0 & 84.3 & 81.5 & 85.2 & 82.3 \\
    NoUnmask &         62.4 & 74.9 & 80.4 & 75.3 & 74.2 & 74.0 & 80.5 & 79.4 & 75.1 \\
    \midrule
    Our &              72.9 & 83.4 & 91.6 & 82.4 & 84.0 & 85.5 & 82.5 & 86.3 & 83.5 \\
    \midrule
    Oracle &           84.4 & 85.2 & 91.6 & 86.1 & 86.0 & 86.5 & 85.3 & 86.5 & 86.4 \\
    \bottomrule
    \end{tabular}
}
\caption{\footnotesize Adversarial Domain Adaptation (ADA) results on the \cite{Blitzer2006DomainAW} dataset. 
}
\label{tab:sentiment_ada}
\end{table*}

%% file: mantis-result.tex
\begin{table}[t]
\centering
\resizebox{0.9\linewidth}{!}{
\begin{tabular}{@{}llllllll@{}}
\toprule
\textbf{Model} & \textbf{AP} & \textbf{DB} & \textbf{EL} & \textbf{PH} & \textbf{ST} & \textbf{UB} & \textit{\textbf{AVG}} \\ \midrule
NoDA       & 71.1 & 67.1 & 72.0 & 47.3 & 66.0 & 72.8 & 66.1 \\
DoCoGen    & 69.8 & 73.9 & 70.7 & 62.0 & 67.8 & 73.4 & 70.1 \\
ReMask     & 70.7 & 73.2 & 72.7 & 60.3 & 68.1 & 73.4 & 70.9 \\
Oracle-Gen & 77.1 & 75.0 & 76.2 & 73.6 & 70.1 & 72.3 & 74.1 \\ \bottomrule
\end{tabular}
}
\caption{\footnotesize UDA results on the MANtIS dataset \shortcite{Penha2019IntroducingMA}. 
}
\label{tab:mantis}
\end{table}